\pdfoutput=1

\documentclass[11pt]{article}

\usepackage{acl}

\usepackage{times}
\usepackage{latexsym}

\usepackage[T1]{fontenc}

\usepackage[utf8]{inputenc}

\usepackage{microtype}
\usepackage{graphicx}

\usepackage{multirow}
\usepackage{color}
\usepackage{colortbl}

\title{How Conservative are Language Models? \\
Adapting to the Introduction of Gender-Neutral Pronouns}

\author{First Author \\
  Affiliation / Address line 1 \\
  Affiliation / Address line 2 \\
  Affiliation / Address line 3 \\
  \texttt{email@domain} \\\And
  Second Author \\
  Affiliation / Address line 1 \\
  Affiliation / Address line 2 \\
  Affiliation / Address line 3 \\
  \texttt{email@domain} \\}
  
\author{Stephanie Brandl, Ruixiang Cui, Anders Søgaard \\
  University of Copenhagen, Denmark\\
  \texttt{\{brandl, rc, soegaard\}@di.ku.dk} 
  }

\begin{document}
\maketitle
\begin{abstract}
Gender-neutral pronouns have recently been introduced in many languages to a) include non-binary people and b) as a generic singular. Recent results from psycholinguistics suggest that gender-neutral pronouns (in Swedish) are {\em not}~associated with human processing difficulties. This, we show, is in sharp contrast with automated processing. We show that gender-neutral pronouns in Danish, English, and Swedish are associated with higher perplexity, more dispersed attention patterns, and worse downstream performance. We argue that such conservativity in language models may limit widespread adoption of gender-neutral pronouns and must therefore be resolved. 
\end{abstract}

\section{Introduction}

Many linguistic scholars have observed how technology in general has altered the course of language evolution \cite{Kristiansen2011LanguageCA,article}, e.g., through the influence of social media conventions. Language technologies, in particular, have also been argued to have such effects, e.g., by reducing the pressure to acquire multiple languages. 

Gender-neutral pronouns is not an entirely modern concept. In 1912, Ella Flag Young, then superintendent of the Chicago public-school system, said the following to a room full of school principals: "The English language is in need of a personal pronoun of the third person, singular number, that will indicate both sexes and will thus eliminate our present awkwardness of speech." The use of gender-neutral pronouns has become much more popular in recent years \cite{gustafsson2021pronouns}. In 2013, a gender-neutral pronoun was {\em politically} introduced in Swedish \cite{gustafsson2015introducing} which can be used for both, people identifying outside the gender dichotomy and as a generic pronoun where information about gender is either unavailable or irrelevant.

In a recently recorded eye-tracking study, \citet{vergoossen2020new} found no evidence that native speakers of Swedish find it harder to process gender-neutral pronouns than gendered pronouns, an argument often brought up by opponents of gender-inclusive language \cite{speyer2019processing, vergoossen2020four}. In combination with their increasing popularity, this suggests gender-neutral pronouns have been or will be widely and fully adapted over time \cite{gustafsson2015introducing, gustafsson2021pronouns}. However, since language technology has the potential to alter the course of language evolution, we want to make sure that our NLP models do not become a bottleneck for this positive development.  

\paragraph{Contribution}
We extract stimuli from a Swedish eye-tracking study that has shown no increase in processing cost in humans for the gender-neutral pronoun \textit{hen} compared to gendered pronouns. We translate those stimuli into English and Danish and  compare model perplexity across gendered and gender-neutral pronouns for all three languages. Furthermore, we systematically investigate performance differences across pronouns in downstream tasks, namely natural language inference (NLI) and coreference resolution. Across the board, we find that NLP models, unlike humans, are challenged by gender-neutral pronouns, incurring significantly higher losses when gendered pronouns are replaced with their gender-neutral alternatives. We argue this is a problem the NLP community must take seriously.\footnote{Our code is available at \href{https://github.com/stephaniebrandl/gender-neutral-pronouns}{\nolinkurl{github.com/stephaniebrandl/gender-neutral-pronouns}}}

\section{Model perplexity and attention}
In this section we introduce a Swedish eye-tracking study and explain how we adapt this study to investigate gender-neutral pronouns in language models.

\begin{table*}[t]
    \centering
   \begin{tabular}{c|c|c|c|c|c|c|c|c|c}
    & \multicolumn{3}{c|}{en} & \multicolumn{3}{c|}{da} & \multicolumn{2}{c}{sv}\\
      & she/he & they & xe & hun/han & de & høn & hon/han & hen\\ \hline
      perplexity & $1 $ & $1.49$ & $2.37$ & $1$ & $1.21$ & $3.35$ & $1$ & $1.8$\\ \hline
      \multirow{3}{*}{correlation} &0.12 &0.26 &0.32 & -0.14 &0.03 & -0.1 &0.19 &0.09\\
      &0.28 &0.33 &0.49 &0.13 &0.17 &0.21 &0.65 &0.72\\
      &0.28 &0.33 &0.49 &0.13 &0.17 &0.22 &0.65 &0.72
    \end{tabular}
    \caption{Perplexity scores across pronouns and languages for the eye-tracking stimuli. Correlation between attention flow and perplexity are listed row-wise for layers 1, 6 and 12.}
    \label{tab:perplexity}
\end{table*}
\paragraph{Humans and \emph{hen}}
\citet{vergoossen2020new} recently recorded a Swedish eye-tracking study to test the hypothesis whether the gender neutral pronoun \textit{hen} has a higher processing cost during pronoun resolution than gendered pronouns. Participants were reading sentence pairs where the first sentence contained a noun referring to a person and the second sentence contained a pronoun referring to that person either with a gendered pronoun or \textit{hen}, for example:
\begin{center}
    \emph{70-åringen dammsög golvet i vardagsrummet. Han/Hen skulle få besök på kvällen.}\\[.5ex]
    \emph{The 70-year-old vacuumed the living room floor. He/They would have visitors in the evening.}
\end{center}

It has recently been shown that attention flow, in contrast to attention itself, correlates with human fixation patterns in task-specific reading \cite{eberle2022large}. We applied a similar analysis pipeline here and extracted all 384 sentence pairs and fed them into the uncased Swedish BERT model.\footnote{\href{https://huggingface.co/af-ai-center/bert-base-swedish-uncased/tree/main}{\nolinkurl{huggingface.co/af-ai-center/bert-base-swedish-uncased}}} We calculate perplexity values for each sentence pair over word probabilities as given by BERT with the formula proposed by \citet{wang-etal-2019-make}. Furthermore, we calculate attention flow \cite{abnar-zuidema-2020-quantifying} propagated from layers 1, 6 and 12 and extract attention flow values assigned to the pronoun with respect to the entity. Attention flow considers the attention matrices as a graph, where tokens are represented as nodes and attention scores as edges between consecutive layers. The edge values, i.e., attention scores, define the maximal flow possible between a pair of nodes. 

We consider different parameters of human fixation which we assume might be influenced by a change in pronouns, in particular during pronoun resolution, i.e., first and total fixation time on the pronoun and fixation time after the first fixation on the noun. For both attention flow and perplexity, however we could not find any meaningful correlation to those parameters. One reason for that might be that the dataset only contains fixations for the two entities, i.e., pronoun and noun, which makes data comparably sparse and impossible to extract complete reading patterns. 
\paragraph{Language models and gender-neutral pronouns} We therefore focus on the model-based data alone in order to understand how well language models can deal with gender-neutral pronouns. For this, we consider perplexity values on sentence-level and calculate rank-based Spearman correlation between perplexity and attention flow for the aforementioned layers. Perplexity has been treated as an indicator for model surprisal and language model quality \cite{goodkind-bicknell-2018-predictive} thus we argue that it serves as a reasonable indicator for processing difficulty. 

With this analysis, we can see if a) gender-neutral pronouns cause a higher sentence perplexity, i.e., a higher \emph{surprisal} and if b) a possible higher surprisal is connected to higher attention flow values on the pronoun with respect to the entity. 

We furthermore translate the sentence pairs into English and Danish where we use two sets of gender-neutral pronouns: 3rd person plural (hence: they/de) which are used in both languages as gender-neutral pronouns \cite{miltersen2020singular} and \emph{neopronouns} (xe for English \cite{hekanaho2020} and høn for Danish).\footnote{\href{https://www.information.dk/kultur/2020/03/hen-hoen-saadan-kom-nye-pronominer-debatten-sproget}{\nolinkurl{information.dk/kultur/hen-hoen}}} For the translation, we use the Google Translate API for Python and manually correct sentences such that semantics agree with the original sentences in Swedish.
We apply the same experiments to those translated datasets with uncased Danish BERT\footnote{\href{https://huggingface.co/Maltehb/danish-bert-botxo}{\nolinkurl{huggingface.co/Maltehb/danish-bert-botxo}}} and uncased English BERT\footnote{\href{https://huggingface.co/bert-base-uncased}{\nolinkurl{ huggingface.co/bert-base-uncased}}}.
\paragraph{Results} We show results on perplexity and correlations in Table \ref{tab:perplexity} for Danish, English and Swedish. Perplexity values for the datasets with gendered pronouns are set to $1$ and we show relative increase for gender-neutral pronouns within a language since perplexity values have been shown to not be comparable across languages \cite{mielke-etal-2019-kind, roh2020unigram}. There we can see that perplexity scores for sentences with gender-neutral pronouns are significantly higher (Wilcoxon signed-rank test resulted in p-values $<0.01$ for all pair-wise comparisons). 

For the correlation between perplexity and attention flow on the Swedish sentence pairs, we can see a clear development between the first layer where there is no correlation ($p>0.05$) for gender-neutral \textit{hen} and very low correlation for gendered pronouns which changes for the other layers where correlations for \textit{hen} are even higher ($\rho=0.72$) than for gendered pronouns ($\rho=0.65$). This suggests that there is some development across layers that is stronger for \textit{hen} than for gendered pronouns. Furthermore, we see a similar evolvement for correlations across layers in English but a much weaker correlation for Danish. 

\begin{figure}
    \centering
    \includegraphics[width=.5\textwidth]{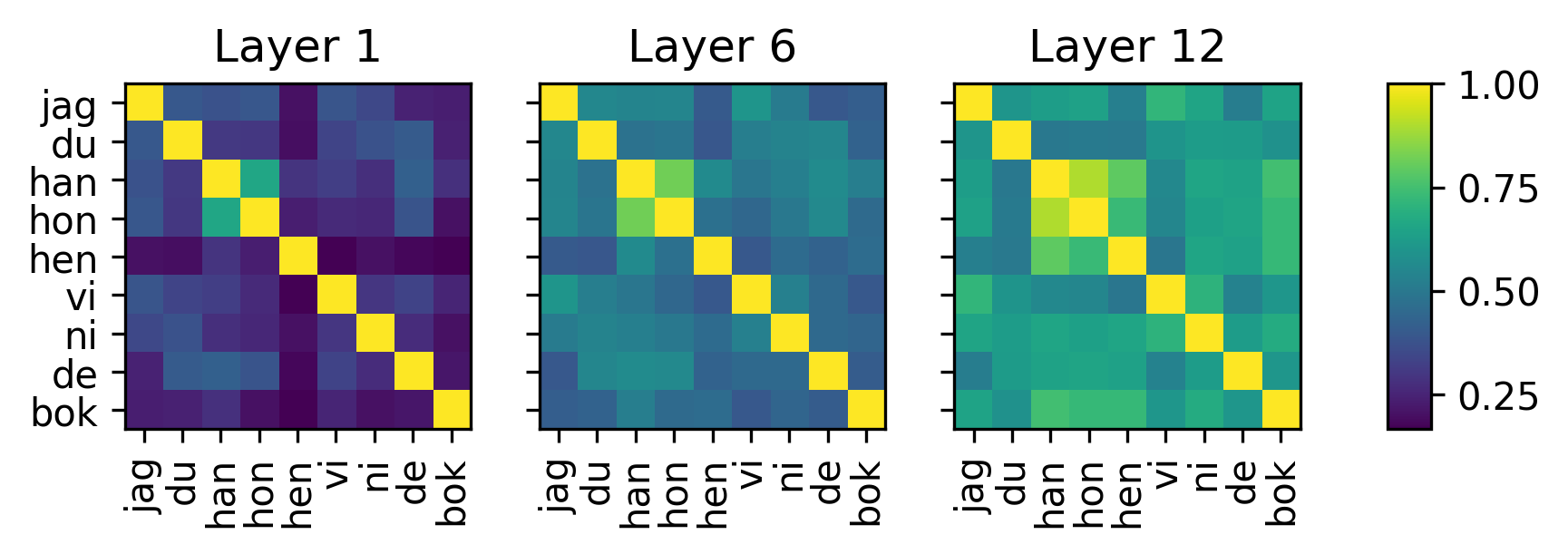}
    \caption{Pair-wise cosine similarity between word representations of all pronouns and the Swedish word \textit{bok} (book) as a baseline for different layers of BERT. We see that gender-neutral \emph{hen} grows from being an outsider (similar to \emph{bok}) in the 1st layer into the cluster of gendered 3rd person pronouns \emph{hon/han} across layers.}
    \label{fig:comparison}
\end{figure}

To investigate those effects across layers further, we look at word embeddings for all Swedish pronouns from all 12 layers in BERT and compute pair-wise cosine similarity including the Swedish word for book (\textit{bok}) as a baseline where we expect no specific relation to pronouns. In Figure \ref{fig:comparison}, we see less similarity between \textit{hen} and the other pronouns in the first layer. This changes for layer 6 and 12 where word representations seem to be more similar and the three 3rd person pronouns \textit{hen, han, hon} get closer to each other. This is in line with the literature where it has been found that single attention heads perform better on pronoun resolution than others. In particular middle and deeper layers have shown stronger attention weights between coreferential elements \cite{vaswani2017attention, webster-etal-2018-mind, clark-etal-2019-bert}. Given that we do not consider individual heads or layers but the entire attention graph it is not surprising that we also see those effects in the top layer as has  been shown in the original paper \cite{abnar-zuidema-2020-quantifying}.

\begin{table*}[t]
    \centering
    \begin{tabular}{c|c|c|c|c|c|c|c|c|c}
    & \multicolumn{3}{c|}{en} & \multicolumn{3}{c|}{da} & \multicolumn{3}{c}{sv}\\
      & orig. & they & xe & orig. & de & høn & orig. & de & hen\\ \hline
      mBERT  & \textbf{83.33} & 83.23 & 81.82 & 71.15 & \textbf{71.24} & 69.72 & \textbf{71.91} & 71.14 & 71.06\\ \hline
      XLM-R & \textbf{95.13} & 94.81 & 94.05 & \textbf{80.19} & 79.18 & 75.48 & \textbf{78.79} & 78.5 & 78.58
    \end{tabular}
    \caption{Accuracy [in \%] on NLI for English, Danish and Swedish for both models mBERT and XLM-R. Accuracies are calculated on the subset of sentences that contain relevant pronouns (924 for en and 2339 for da/sv). The first column for each language shows the accuracy on the original data, second and third columns show accuracies for respective gender-neutral pronouns. Please note, the total number of label flips in both directions for different pronouns is higher than the performance difference for all pair-wise comparisons. A baseline analysis where we exchanged punctuation ("." for "!") yields similar deviations from the original dataset than the changing pronouns.}
    \label{tab:nli}
\end{table*}

\section{Downstream Tasks}
We also perform downstream task experiments on natural language inference and coreference resolution for both gendered and gender-neutral pronouns to investigate to what extent gender-neutral pronouns influence the performance.

\paragraph{Natural Language Inference}
Natural Language Inference (NLI) is commonly framed as a classification task, which tests a model's ability to understand entailment and contradiction \citep{bowman-etal-2015-large}. Despite high accuracies achieved by SOTA models, we are yet to know whether they succeed in combating gender bias, especially in cross-lingual settings. We apply two multilingual models mBERT\footnote{\texttt{multi\_cased\_L-12\_H-768\_A-12}} \citep{devlin-etal-2019-bert} and XLM-R\footnote{\texttt{xlm-roberta-large}} \citep{conneau-etal-2020-unsupervised} with cross-lingual fine-tuning, i.e., we fine-tune on English and apply both models also on Danish and Swedish. Therefore, mBERT was fine-tuned on the English MNLI train split and evaluated on XNLI. For XLM-R, we apply a model that has been fine-tuned on both MNLI and ANLI \citep{nie-etal-2020-adversarial}\footnote{\href{https://huggingface.co/vicgalle/xlm-roberta-large-xnli-anli}{\nolinkurl{huggingface.co/vicgalle/xlm-roberta-large-xnli-anli}}}. For English we test both models on the MNLI test split, for Danish and Swedish we test on the extended XNLI corpus \cite{singhMultiLingTokBias2019}, the manual translation of the first 15000 sentences of the MNLI corpus \citep{williams-etal-2018-broad} from English into 15 languages.

\paragraph{Coreference Resolution}
We also run pronoun resolution experiments on the Winogender dataset \cite{rudinger-etal-2018-gender} where all 720 English sentences include an \textit{occupation}, a \textit{participant} and a \textit{pronoun}. For each occupation, two similar sentences are composed, one where the pronoun refers to the occupation and one where it refers to the participant. Those sentences are then presented in versions with different pronouns (female, male, singular they). For our experiments, we compare performance for those pronouns and add a version for the gender-neutral pronoun \textit{xe}. We run experiments with NeuralCoref 4.0 in SpaCy.\footnote{\href{https://github.com/huggingface/neuralcoref}{\nolinkurl{github.com/huggingface/neuralcoref}}}. \citet{lauscher2022welcome} conduct similar experiments in English where all pronouns are exchanged for their POS tag, in contrast to our experiments where we only exchange gendered pronouns and replace them with gender-neutral pronouns. 

For Danish, we apply the recently published coreference model \cite{barrett-etal-2021-resources} to both the corresponding test set from the \textit{Dacoref} dataset and a \textit{gender-neutralized} version where we exchange gendered pronouns \textit{hun/han} for either \textit{høn} or singular \textit{de}.\footnote{So far, no Swedish coreference model has been published, we therefore leave this analysis for future work.}

\section{Results}
\paragraph{Natural Language Inference}
Accuracies for all languages and both models are displayed in Table \ref{tab:nli}. We overall see a very small drop in performance for the datasets with gender neutral pronouns compared to the original sentences. For mBERT we see differences of $0.09-1.51\%$, for XLM-R the drop is slightly higher with $0.21-4.71\%$. We see the biggest difference for the Danish pronoun \textit{høn} in comparison to the original dataset. 

\paragraph{Coreference Resolution}
\begin{table}[h]
    \centering
    \begin{tabular}{c|c|c|c|c}
        & she & he & they & xe\\ \hline
         acc in \% & 42.92 & \textbf{43.75} & 27.92 & 0 
    \end{tabular}
    \caption{Results for the pronoun resolution task on the English Winogender dataset.}
    \label{tab:coref}
\end{table}

\begin{table}[h]
    \centering
    \begin{tabular}{r|c|c|c}
         & orig. & de & høn \\ \hline
         F1-score & \textbf{0.64} & 0.63 & 0.62\\
         Prec. & \textbf{0.70} & 0.69 & 0.69 \\
         Recall & \textbf{0.59} & 0.57 & 0.56
    \end{tabular}
    \caption{Results for the Danish coreference resolution task. Pronouns in the original dataset (orig.) have been exchanged for singular \textit{de} and gender-neutral \textit{høn}.}
    \label{tab:coref_da}
\end{table}
Table \ref{tab:coref} shows accuracies on the English Winogender corpus for all four pronouns. We see a clear drop in performance from gendered pronouns (\textit{she, he}) to both gender-neutral pronouns (\textit{they, xe}). For \textit{xe}, the model was not able to perform coreference resolution at all. In most cases it was not even recognized as part of a cluster and in the rare cases where it was, it was clustered with the wrong tokens. Please note that since this dataset is not labelled we are only classifying if the pronoun has been clustered with the correct entity. 

Results on the Danish Coref corpus, where we are able to perform a more extensive coreference resolution task are displayed in Table \ref{tab:coref_da}. We were able to replicate results from \citet{barrett-etal-2021-resources} (the first column \textit{orig.}) and see small drops in performance for singular \textit{de} and \textit{høn}.

\label{sec:appendix}
\section{Related Work}
More eye-tracking studies have been conducted investigating the influence in processing cost for both gender-neutral pronouns and the generic male pronoun. \citet{irmen2007s} and \citet{redl2021male} find male biases when using generic male pronouns in Dutch and generic role nouns in German. The authors of \citet{sanford2007they} found a clear processing cost when using singular \textit{they} in English, however their stimuli did not include any investigation of how (anti-)stereotypes influence this processing cost and is thus only in parts comparable to other studies. English datasets have been proposed to investigate gender bias in pronoun resolution but have not reported on performance differences between gendered and gender-neutral pronouns \cite{rudinger-etal-2018-gender, zhao2018gender, webster-etal-2018-mind}. \citet{sun2021they} propose a rewriting task where data is transferred from gendered to gender-neutral pronouns to train more inclusive language models. \citet{cao-daume-iii-2020-toward} and \citet{dev-etal-2021-harms} discuss the necessity of including non-binary pronouns into NLP research (see also \citet{stanczak2021survey}). 

\section{Discussion}
With this paper we provide a first study on how well language can handle gender-neutral pronouns in Danish, English and Swedish for various tasks. We observe an increase in perplexity for gender-neutral pronouns and correlations between perplexity on sentence level and attention flow on the pronoun, in particular for English and Swedish that gets stronger across layers. This indicates that language models indeed struggle with the use of gender-neutral pronouns, even with singular \emph{they}, which has been used for many years as gender-neutral \cite{saguy2022little}. The reason for this most likely lies in the sparse representation of gender-neutral pronouns in the training data and the fact that language models, once they are trained and published usually are not updated \cite{bender2021dangers}. However, Transformer models pre-trained on subword units have been shown to be robust with respect to word frequency \cite{sennrich-etal-2016-neural} and thus should be able to process unfamiliar gender-neutral pronouns. At the same time, we observe that word representations of all Swedish 3rd person pronouns grow closer in middle and top layers (see Figure 1) which suggests that relevant information is also learned for gender-neutral \emph{hen}.

For NLI, we only see a small drop in performance when exchanging gendered pronouns for gender-neutral pronouns which is in the same range as a baseline analysis where we exchange punctuation ("!" for "."), except for Danish \emph{høn}. We argue that classification in NLI probably does not heavily rely on individual pronouns in most cases.  In stark contrast to pronoun resolution where we see a very clear drop in performance for English when applying singular \emph{they} in comparison to both female and male pronouns, again this is surprising since in theory language models should have seen training samples where singular \emph{they} has been used. The small drop in performance for Danish coreference resolution might be because this dataset does not solely focus on pronoun resolution, though further investigation is needed here. We strongly argue that more needs to be done to adapt language models to a more gender inclusive language, initiatives like the rewriting task as proposed by \citet{sun2021they} need to be implemented and extended.

\section*{Acknowledgements}
This  work  was partially funded by the Platform Intelligence in News project, which is supported by Innovation Fund Denmark via the Grand Solutions program. We thank Vinit Ravishankar and Jonas Lotz for fruitful discussions and Daniel Hershcovich and Yova Kementchedjhieva for proof-reading and valuable inputs on the manuscript. We also thank Kellie Webster for her valuable input on gender-neutral pronouns.

\bibliography{anthology,custom}

\begin{thebibliography}{38}
\expandafter\ifx\csname natexlab\endcsname\relax\def\natexlab#1{#1}\fi

\bibitem[{Abbasi(2020)}]{article}
Irum Abbasi. 2020.
\newblock \href {https://doi.org/10.5539/elt.v13n7p1} {The influence of
  technology on english language and literature}.
\newblock \emph{English Language Teaching}, 13:1.

\bibitem[{Abnar and Zuidema(2020)}]{abnar-zuidema-2020-quantifying}
Samira Abnar and Willem Zuidema. 2020.
\newblock \href {https://doi.org/10.18653/v1/2020.acl-main.385} {Quantifying
  attention flow in transformers}.
\newblock In \emph{Proceedings of the 58th Annual Meeting of the Association
  for Computational Linguistics}, pages 4190--4197, Online. Association for
  Computational Linguistics.

\bibitem[{Barrett et~al.(2021)Barrett, Lam, Wu, Lacroix, Plank, and
  S{\o}gaard}]{barrett-etal-2021-resources}
Maria Barrett, Hieu Lam, Martin Wu, Oph{\'e}lie Lacroix, Barbara Plank, and
  Anders S{\o}gaard. 2021.
\newblock \href {https://doi.org/10.18653/v1/2021.crac-1.7} {Resources and
  evaluations for {D}anish entity resolution}.
\newblock In \emph{Proceedings of the Fourth Workshop on Computational Models
  of Reference, Anaphora and Coreference}, pages 63--69, Punta Cana, Dominican
  Republic. Association for Computational Linguistics.

\bibitem[{Bender et~al.(2021)Bender, Gebru, McMillan-Major, and
  Shmitchell}]{bender2021dangers}
Emily~M Bender, Timnit Gebru, Angelina McMillan-Major, and Shmargaret
  Shmitchell. 2021.
\newblock On the dangers of stochastic parrots: Can language models be too big?
\newblock In \emph{Proceedings of the 2021 ACM Conference on Fairness,
  Accountability, and Transparency}, pages 610--623.

\bibitem[{Bowman et~al.(2015)Bowman, Angeli, Potts, and
  Manning}]{bowman-etal-2015-large}
Samuel~R. Bowman, Gabor Angeli, Christopher Potts, and Christopher~D. Manning.
  2015.
\newblock \href {https://doi.org/10.18653/v1/D15-1075} {A large annotated
  corpus for learning natural language inference}.
\newblock In \emph{Proceedings of the 2015 Conference on Empirical Methods in
  Natural Language Processing}, pages 632--642, Lisbon, Portugal. Association
  for Computational Linguistics.

\bibitem[{Cao and Daum{\'e}~III(2020)}]{cao-daume-iii-2020-toward}
Yang~Trista Cao and Hal Daum{\'e}~III. 2020.
\newblock \href {https://doi.org/10.18653/v1/2020.acl-main.418} {Toward
  gender-inclusive coreference resolution}.
\newblock In \emph{Proceedings of the 58th Annual Meeting of the Association
  for Computational Linguistics}, pages 4568--4595, Online. Association for
  Computational Linguistics.

\bibitem[{Clark et~al.(2019)Clark, Khandelwal, Levy, and
  Manning}]{clark-etal-2019-bert}
Kevin Clark, Urvashi Khandelwal, Omer Levy, and Christopher~D. Manning. 2019.
\newblock \href {https://doi.org/10.18653/v1/W19-4828} {What does {BERT} look
  at? an analysis of {BERT}{'}s attention}.
\newblock In \emph{Proceedings of the 2019 ACL Workshop BlackboxNLP: Analyzing
  and Interpreting Neural Networks for NLP}, pages 276--286, Florence, Italy.
  Association for Computational Linguistics.

\bibitem[{Conneau et~al.(2020)Conneau, Khandelwal, Goyal, Chaudhary, Wenzek,
  Guzm{\'a}n, Grave, Ott, Zettlemoyer, and
  Stoyanov}]{conneau-etal-2020-unsupervised}
Alexis Conneau, Kartikay Khandelwal, Naman Goyal, Vishrav Chaudhary, Guillaume
  Wenzek, Francisco Guzm{\'a}n, Edouard Grave, Myle Ott, Luke Zettlemoyer, and
  Veselin Stoyanov. 2020.
\newblock \href {https://doi.org/10.18653/v1/2020.acl-main.747} {Unsupervised
  cross-lingual representation learning at scale}.
\newblock In \emph{Proceedings of the 58th Annual Meeting of the Association
  for Computational Linguistics}, pages 8440--8451, Online. Association for
  Computational Linguistics.

\bibitem[{Dev et~al.(2021)Dev, Monajatipoor, Ovalle, Subramonian, Phillips, and
  Chang}]{dev-etal-2021-harms}
Sunipa Dev, Masoud Monajatipoor, Anaelia Ovalle, Arjun Subramonian, Jeff
  Phillips, and Kai-Wei Chang. 2021.
\newblock \href {https://doi.org/10.18653/v1/2021.emnlp-main.150} {Harms of
  gender exclusivity and challenges in non-binary representation in language
  technologies}.
\newblock In \emph{Proceedings of the 2021 Conference on Empirical Methods in
  Natural Language Processing}, pages 1968--1994, Online and Punta Cana,
  Dominican Republic. Association for Computational Linguistics.

\bibitem[{Devlin et~al.(2019)Devlin, Chang, Lee, and
  Toutanova}]{devlin-etal-2019-bert}
Jacob Devlin, Ming-Wei Chang, Kenton Lee, and Kristina Toutanova. 2019.
\newblock \href {https://doi.org/10.18653/v1/N19-1423} {{BERT}: Pre-training of
  deep bidirectional transformers for language understanding}.
\newblock In \emph{Proceedings of the 2019 Conference of the North {A}merican
  Chapter of the Association for Computational Linguistics: Human Language
  Technologies, Volume 1 (Long and Short Papers)}, pages 4171--4186,
  Minneapolis, Minnesota. Association for Computational Linguistics.

\bibitem[{Eberle et~al.(2022)Eberle, Brandl, Pilot, and
  Søgaard}]{eberle2022large}
Oliver Eberle, Stephanie Brandl, Jonas Pilot, and Anders Søgaard. 2022.
\newblock \href {https://openreview.net/forum?id=PjeQeQtWF36} {Do transformer
  models show similar attention patterns to task-specific human gaze?}
\newblock To appear at ACL 2022.

\bibitem[{Goodkind and Bicknell(2018)}]{goodkind-bicknell-2018-predictive}
Adam Goodkind and Klinton Bicknell. 2018.
\newblock \href {https://doi.org/10.18653/v1/W18-0102} {Predictive power of
  word surprisal for reading times is a linear function of language model
  quality}.
\newblock In \emph{Proceedings of the 8th Workshop on Cognitive Modeling and
  Computational Linguistics ({CMCL} 2018)}, pages 10--18, Salt Lake City, Utah.
  Association for Computational Linguistics.

\bibitem[{Gustafsson~Send{\'e}n et~al.(2015)Gustafsson~Send{\'e}n, B{\"a}ck,
  and Lindqvist}]{gustafsson2015introducing}
Marie Gustafsson~Send{\'e}n, Emma~A B{\"a}ck, and Anna Lindqvist. 2015.
\newblock Introducing a gender-neutral pronoun in a natural gender language:
  the influence of time on attitudes and behavior.
\newblock \emph{Frontiers in Psychology}, 6:893.

\bibitem[{Gustafsson~Send{\'e}n et~al.(2021)Gustafsson~Send{\'e}n,
  Renstr{\"o}m, and Lindqvist}]{gustafsson2021pronouns}
Marie Gustafsson~Send{\'e}n, Emma Renstr{\"o}m, and Anna Lindqvist. 2021.
\newblock Pronouns beyond the binary: The change of attitudes and use over
  time.
\newblock \emph{Gender \& Society}, 35(4):588--615.

\bibitem[{Hekanaho(2020)}]{hekanaho2020}
Laura Hekanaho. 2020.
\newblock \emph{Generic and Nonbinary Pronouns : Usage, Acceptability and
  Attitudes}.
\newblock Ph.D. thesis, University of Helsink, Helsinki, Finland.

\bibitem[{Irmen(2007)}]{irmen2007s}
Lisa Irmen. 2007.
\newblock What’s in a (role) name? formal and conceptual aspects of
  comprehending personal nouns.
\newblock \emph{Journal of Psycholinguistic Research}, 36(6):431--456.

\bibitem[{Kristiansen et~al.(2011)Kristiansen, Coupland, Soukup,
  Moosm{\"u}ller, Gregersen, Garrett, Selleck, Nuolij{\"a}rvi, Vaattovaara,
  {\"O}stman, Mattfolk, Stoeckle, Svenstrup, Leonard, {\'A}rnason,
  Hifearn{\'a}in, Murchadha, Vaicekauskienė, Grondelaers, van Hout, Sand{\o}y,
  Thelander, Robert, Androutsopoulos, Auer, Spiekermann, Bell, Speelman, and
  Stuart-Smith}]{Kristiansen2011LanguageCA}
Tore Kristiansen, Nikolas Coupland, Barbara Soukup, Sylvia Moosm{\"u}ller,
  Frans Gregersen, Peter Garrett, Charlotte Selleck, Pirkko Nuolij{\"a}rvi,
  Johanna Vaattovaara, Jan-Ola~Ingemar {\"O}stman, Leila Mattfolk, Philipp
  Stoeckle, Christoph~Hare Svenstrup, Stephen~Pax Leonard, Kristj{\'a}n
  {\'A}rnason, Tadhg~{\'O} Hifearn{\'a}in, Noel~{\'O} Murchadha, Loreta
  Vaicekauskienė, Stefan Grondelaers, Roeland van Hout, Helge Sand{\o}y, Mats
  Thelander, Elen Robert, Jannis Androutsopoulos, Peter Auer, Helmut
  Spiekermann, Allan Bell, Dirk Speelman, and Jane Stuart-Smith. 2011.
\newblock Language change and digital media: A review of conceptions and
  evidence.
\newblock In \emph{Standard languages and language standards in a changing
  Europe}.

\bibitem[{Lauscher et~al.(2022)Lauscher, Crowley, and
  Hovy}]{lauscher2022welcome}
Anne Lauscher, Archie Crowley, and Dirk Hovy. 2022.
\newblock Welcome to the modern world of pronouns: Identity-inclusive natural
  language processing beyond gender.
\newblock \emph{arXiv preprint arXiv:2202.11923}.

\bibitem[{Mielke et~al.(2019)Mielke, Cotterell, Gorman, Roark, and
  Eisner}]{mielke-etal-2019-kind}
Sabrina~J. Mielke, Ryan Cotterell, Kyle Gorman, Brian Roark, and Jason Eisner.
  2019.
\newblock \href {https://doi.org/10.18653/v1/P19-1491} {What kind of language
  is hard to language-model?}
\newblock In \emph{Proceedings of the 57th Annual Meeting of the Association
  for Computational Linguistics}, pages 4975--4989, Florence, Italy.
  Association for Computational Linguistics.

\bibitem[{Miltersen(2020)}]{miltersen2020singular}
Ehm~Hjorth Miltersen. 2020.
\newblock Singular de and its referential use in talk-in-interaction.
\newblock \emph{Scandinavian Studies in Language}, 11(2):37--37.

\bibitem[{Nie et~al.(2020)Nie, Williams, Dinan, Bansal, Weston, and
  Kiela}]{nie-etal-2020-adversarial}
Yixin Nie, Adina Williams, Emily Dinan, Mohit Bansal, Jason Weston, and Douwe
  Kiela. 2020.
\newblock \href {https://doi.org/10.18653/v1/2020.acl-main.441} {Adversarial
  {NLI}: A new benchmark for natural language understanding}.
\newblock In \emph{Proceedings of the 58th Annual Meeting of the Association
  for Computational Linguistics}, pages 4885--4901, Online. Association for
  Computational Linguistics.

\bibitem[{Redl et~al.(2021)Redl, Frank, De~Swart, and De~Hoop}]{redl2021male}
Theresa Redl, Stefan~L Frank, Peter De~Swart, and Helen De~Hoop. 2021.
\newblock The male bias of a generically-intended masculine pronoun: Evidence
  from eye-tracking and sentence evaluation.
\newblock \emph{PloS one}, 16(4):e0249309.

\bibitem[{Roh et~al.(2020)Roh, Oh, and Lee}]{roh2020unigram}
Jihyeon Roh, Sang-Hoon Oh, and Soo-Young Lee. 2020.
\newblock Unigram-normalized perplexity as a language model performance measure
  with different vocabulary sizes.
\newblock \emph{arXiv preprint arXiv:2011.13220}.

\bibitem[{Rudinger et~al.(2018)Rudinger, Naradowsky, Leonard, and
  Van~Durme}]{rudinger-etal-2018-gender}
Rachel Rudinger, Jason Naradowsky, Brian Leonard, and Benjamin Van~Durme. 2018.
\newblock \href {https://doi.org/10.18653/v1/N18-2002} {Gender bias in
  coreference resolution}.
\newblock In \emph{Proceedings of the 2018 Conference of the North {A}merican
  Chapter of the Association for Computational Linguistics: Human Language
  Technologies, Volume 2 (Short Papers)}, pages 8--14, New Orleans, Louisiana.
  Association for Computational Linguistics.

\bibitem[{Saguy and Williams(2022)}]{saguy2022little}
Abigail~C Saguy and Juliet~A Williams. 2022.
\newblock A little word that means a lot: A reassessment of singular they in a
  new era of gender politics.
\newblock \emph{Gender \& Society}, 36(1):5--31.

\bibitem[{Sanford and Filik(2007)}]{sanford2007they}
Anthony~J Sanford and Ruth Filik. 2007.
\newblock “they” as a gender-unspecified singular pronoun: Eye tracking
  reveals a processing cost.
\newblock \emph{Quarterly Journal of Experimental Psychology}, 60(2):171--178.

\bibitem[{Sennrich et~al.(2016)Sennrich, Haddow, and
  Birch}]{sennrich-etal-2016-neural}
Rico Sennrich, Barry Haddow, and Alexandra Birch. 2016.
\newblock \href {https://doi.org/10.18653/v1/P16-1162} {Neural machine
  translation of rare words with subword units}.
\newblock In \emph{Proceedings of the 54th Annual Meeting of the Association
  for Computational Linguistics (Volume 1: Long Papers)}, pages 1715--1725,
  Berlin, Germany. Association for Computational Linguistics.

\bibitem[{Singh et~al.(2019)Singh, McCann, Xiong, and
  Socher}]{singhMultiLingTokBias2019}
Jasdeep Singh, Bryan McCann, Caiming Xiong, and Richard Socher. 2019.
\newblock {BERT is Not an Interlingua and the Bias of Tokenization}.
\newblock \emph{The Workshop on Deep Learning for Low-Resource NLP at EMNLP
  2019}.

\bibitem[{Speyer and Schleef(2019)}]{speyer2019processing}
Lydia~Gabriela Speyer and Erik Schleef. 2019.
\newblock Processing ‘gender-neutral’pronouns: A self-paced reading study
  of learners of english.
\newblock \emph{Applied Linguistics}, 40(5):793--815.

\bibitem[{Stanczak and Augenstein(2021)}]{stanczak2021survey}
Karolina Stanczak and Isabelle Augenstein. 2021.
\newblock A survey on gender bias in natural language processing.
\newblock \emph{arXiv preprint arXiv:2112.14168}.

\bibitem[{Sun et~al.(2021)Sun, Webster, Shah, Wang, and Johnson}]{sun2021they}
Tony Sun, Kellie Webster, Apu Shah, William~Yang Wang, and Melvin Johnson.
  2021.
\newblock They, them, theirs: Rewriting with gender-neutral english.
\newblock \emph{arXiv preprint arXiv:2102.06788}.

\bibitem[{Vaswani et~al.(2017)Vaswani, Shazeer, Parmar, Uszkoreit, Jones,
  Gomez, Kaiser, and Polosukhin}]{vaswani2017attention}
Ashish Vaswani, Noam Shazeer, Niki Parmar, Jakob Uszkoreit, Llion Jones,
  Aidan~N Gomez, Lukasz Kaiser, and Illia Polosukhin. 2017.
\newblock Attention is all you need.
\newblock \emph{arXiv preprint arXiv:1706.03762}.

\bibitem[{Vergoossen et~al.(2020{\natexlab{a}})Vergoossen, P{\"a}rnamets,
  Renstr{\"o}m, and Gustafsson~Send{\'e}n}]{vergoossen2020new}
Hellen~P Vergoossen, Philip P{\"a}rnamets, Emma~A Renstr{\"o}m, and Marie
  Gustafsson~Send{\'e}n. 2020{\natexlab{a}}.
\newblock Are new gender-neutral pronouns difficult to process in reading? the
  case of hen in swedish.
\newblock \emph{Frontiers in psychology}, 11:2967.

\bibitem[{Vergoossen et~al.(2020{\natexlab{b}})Vergoossen, Renstr{\"o}m,
  Lindqvist, and Send{\'e}n}]{vergoossen2020four}
Hellen~Petronella Vergoossen, Emma~Aurora Renstr{\"o}m, Anna Lindqvist, and
  Marie~Gustafsson Send{\'e}n. 2020{\natexlab{b}}.
\newblock Four dimensions of criticism against gender-fair language.
\newblock \emph{Sex Roles}, 83(5):328--337.

\bibitem[{Wang et~al.(2019)Wang, Liang, Zhang, Li, and
  Gao}]{wang-etal-2019-make}
Cunxiang Wang, Shuailong Liang, Yue Zhang, Xiaonan Li, and Tian Gao. 2019.
\newblock \href {https://doi.org/10.18653/v1/P19-1393} {Does it make sense? and
  why? a pilot study for sense making and explanation}.
\newblock In \emph{Proceedings of the 57th Annual Meeting of the Association
  for Computational Linguistics}, pages 4020--4026, Florence, Italy.
  Association for Computational Linguistics.

\bibitem[{Webster et~al.(2018)Webster, Recasens, Axelrod, and
  Baldridge}]{webster-etal-2018-mind}
Kellie Webster, Marta Recasens, Vera Axelrod, and Jason Baldridge. 2018.
\newblock \href {https://doi.org/10.1162/tacl_a_00240} {Mind the {GAP}: A
  balanced corpus of gendered ambiguous pronouns}.
\newblock \emph{Transactions of the Association for Computational Linguistics},
  6:605--617.

\bibitem[{Williams et~al.(2018)Williams, Nangia, and
  Bowman}]{williams-etal-2018-broad}
Adina Williams, Nikita Nangia, and Samuel Bowman. 2018.
\newblock \href {https://doi.org/10.18653/v1/N18-1101} {A broad-coverage
  challenge corpus for sentence understanding through inference}.
\newblock In \emph{Proceedings of the 2018 Conference of the North {A}merican
  Chapter of the Association for Computational Linguistics: Human Language
  Technologies, Volume 1 (Long Papers)}, pages 1112--1122, New Orleans,
  Louisiana. Association for Computational Linguistics.

\bibitem[{Zhao et~al.(2018)Zhao, Wang, Yatskar, Ordonez, and
  Chang}]{zhao2018gender}
Jieyu Zhao, Tianlu Wang, Mark Yatskar, Vicente Ordonez, and Kai-Wei Chang.
  2018.
\newblock Gender bias in coreference resolution: Evaluation and debiasing
  methods.
\newblock In \emph{Proceedings of the 2018 Conference of the North American
  Chapter of the Association for Computational Linguistics: Human Language
  Technologies, Volume 2 (Short Papers)}, pages 15--20.

\end{thebibliography}
\bibliographystyle{acl_natbib}
\end{document}